\documentclass{esagnc}

\usepackage[
    backend=biber,
    style=ieee,
    citestyle=numeric-comp,
    mincitenames=1,
    maxcitenames=2,
    maxbibnames=5]
           {biblatex}
\usepackage{csquotes}
\usepackage{doi}        
\usepackage{graphicx}
\usepackage{subcaption}
\usepackage[acronym]{glossaries-extra}  
\usepackage{glossary-mcols}             
\usepackage{siunitx}                    

\DeclareNameAlias{author}{family-given}
\DeclareNameAlias{editor}{family-given}
\DeclareNameAlias{translator}{family-given}
\newignoredglossary{hidden}
\glsdisablehyper
\hypersetup{
  linkcolor  = [rgb]{0.3098,0.5059,0.7412},
  citecolor  = [rgb]{0.3098,0.5059,0.7412},
  urlcolor   = [rgb]{0.3098,0.5059,0.7412},
  colorlinks = true,
}
\AtBeginDocument{\sisetup{math-rm=\mathrm, detect-weight=true, binary-units = true, mode=text,range-phrase = --}}
\DeclareSIUnit\degree{deg}
\DeclareSIUnit\usd{USD}
\DeclareSIUnit\rpm{rpm}
\DeclareSIUnit\gauss{G}
\DeclareSIUnit\pixel{px}

\def\glssquare#1{%
\ifglsused{#1}
    {\glsxtrshort{#1}%
    }
    {\glsxtrshort{#1} [\glsxtrlong*{#1}]\glsunset{#1}%
    }%
}

\def\figref#1{Fig.~\ref{#1}}
\def\Figref#1{Figure~\ref{#1}}
\def\Tabref#1{Table~\ref{#1}}
\def\eqref#1{Eq.~(\ref{#1})}

\newacronym[category=hidden]{adr}{ADR}{Active Debris Removal}
\newacronym[category=hidden]{asl}{ASL}{Autonomous Systems Laboratory}
\newacronym[category=hidden]{atv}{ATV}{Automated Transfer Vehicle}

\newacronym[category=hidden]{cad}{CAD}{Computer-Aided Design}
\newacronym[category=hidden]{cnes}{CNES}{Centre National d'Études Spatiales, France}
\newacronym[category=hidden]{cnn}{CNN}{Convolutional Neural Network}

\newacronym[category=hidden,longplural=Degrees-Of-Freedom]{dof}{DOF}{Degree-Of-Freedom}
\newacronym[category=hidden]{dnn}{DNN}{Deep Neural Network}

\newacronym[category=hidden]{gnc}{GNC}{Guidance, Navigation, and Control}

\newacronym[category=hidden]{iss}{ISS}{International Space Station}

\newacronym[category=hidden]{lidar}{LIDAR}{LIght Detection and Ranging}
\newacronym[category=hidden]{lwir}{LWIR}{Long Wavelength Infrared}

\newacronym[category=hidden]{nasa}{NASA}{National Aeronautics and Space Administration, USA}

\title{\uppercase{Using Convolutional Neural Networks for Relative Pose Estimation of a Non-Cooperative Spacecraft with Thermal Infrared Imagery}}
\author[(1)]{Maxwell Hogan}
\author[(1)]{Duarte Rondao}
\author[(1)]{Nabil Aouf}
\author[(2)]{Olivier Dubois-Matra}
\affil[(1)]{City, University of London, ECV1 0HB London, United Kingdom\\\texttt{\{firstname.lastname\}@city.ac.uk}}
\affil[(2)]{European Space Agency, ESTEC, Keplerlaan 1, 2201 AZ Noordwijk, The Netherlands\\\texttt{olivier.dubois-matra@esa.int}}

\footauth{M. Hogan, D. Rondao, N. Aouf, O. Dubois-Matra}

\begin{document}

\maketitle

\begin{abstract}
Recent interest in on-orbit servicing and \gls{adr} missions have driven the need for technologies to enable non-cooperative rendezvous manoeuvres. Such manoeuvres put heavy burden on the perception capabilities of a chaser spacecraft. This paper demonstrates \glspl{cnn} capable of providing an initial coarse pose estimation of a target from a passive thermal infrared camera feed. Thermal cameras offer a promising alternative to visible cameras, which struggle in low light conditions and are susceptible to overexposure. Often, thermal information on the target is not available \textit{a priori}; this paper therefore proposes using visible images to train networks. The robustness of the models is demonstrated on two different targets, first on synthetic data, and then in a laboratory environment for a realistic scenario that might be faced during an \gls{adr} mission. Given that there is much concern over the use of \glspl{cnn} in critical applications due to their black box nature, we use innovative techniques to explain what is important to our network and fault conditions.  

\section*{KEYWORDS}
Deep Learning; Explainable Artificial Intelligence, Computer Vision; Spacecraft Pose Estimation; Active Debris Removal.
\end{abstract}

\glsresetall

\section{INTRODUCTION}

Autonomous rendezvous and docking missions such as the \gls{iss} resupply missions undertaken by the European \gls{atv} and more recently, SpaceX Cargo Dragon have had the advantage of a cooperating target. Markers and additional sensors on the target, and a physical interface to grapple, were used to ensure a safe rendezvous. However, recent interests have shifted to technologies for use in situations where the target is unresponsive such as \gls{adr} missions. The lack of aid from the target puts higher responsibility on the chaser’s \gls{gnc} system. To the authors' knowledge, such a manoeuvre has only been achieved once with the capture of and relaunch of the IntelSat VI satellite which required human intervention for success \cite{fricke1992sts}.

A natural solution might be the use of a \gls{lidar} system which has already seen use on the Space Shuttle. \gls{lidar} sensors can supply range information with high accuracy without being susceptible to illumination changes which can be expected in on orbit conditions \cite{opromolla2017review}. However, they remain difficult to install on small craft due to their size and heavy power requirements. On the other hand, passive sensors such as visible and infrared cameras are characterised by a lower hardware complexity and cheaper power consumption \cite{opromolla2017review}.

It is the motivation of this project to extend the capabilities of passive sensors for close-range relative navigation. Other works have presented the use of \glspl{cnn} to provide pose estimation of non-cooperative spacecraft on visible images; however, these works have proposed their use as a full-fledged coarse estimator \cite{sharma2019pose,proenca2019deep}. \glspl{cnn} do not take into consideration the previous prediction and ignore the temporal relationship between predictions. This can result in high variance between the solutions of contiguous frames.

We have chosen to focus on a simpler problem statement which is to provide an initial estimation of the pose so that a more precise algorithm can make refinements. This had been previously achieved by \textcite{rondao2021robust} where a Bayesian classifier, a traditional machine learning algorithm, was used to match 2D features from a visible camera feed to keyframes of the 3D target that were learned offline. This method does automatically extract feature from an image but these features must be determined by the developer beforehand. This presents a difficulty since a developer would have to consider the effect of a large possible scenarios when they are making their decision. Conversely, the features that are extracted by a \glspl{cnn} are learned through optimisation. A \glspl{cnn} can instead learn generalisations about the data that make it more stable for dealing with fringe cases.

Visible light cameras (\SIrange{0.38}{0.75}{\micro\metre}) struggle to match the reliability of continuous measurements of active sensors due to the poor illumination conditions in eclipse and oversaturation from reflective surfaces or direct light from the sun. One the other hand, recent studies suggest that thermal infrared (or \glssquare{lwir}) based cameras offer a promising alternative to vision-based or active technologies for non-cooperative rendezvous missions \cite{hajebi2008structure,yilmaz2017using}. According to a report by \textcite{yilmaz2017using}, objects under direct illumination from the Sun should appear smoother in \gls{lwir} (\SIrange{8}{14}{\micro\metre}) when compared to visible due to reflective surfaces such as thermal insulators. An additional benefit of \gls{lwir} images is the lack of shadows created by incident light; therefore, these should require less intensive pre-processing to clean the image.

A significant challenge in the development of \gls{lwir}-based \glspl{cnn} is the lack of relevant thermal infrared datasets for training. Nevertheless, visible data of a target is often easily acquired and often already available. Therefore, it is the motivation of this study to investigate the robustness of a \gls{cnn} tested on thermal images which has been trained exclusively on synthetic visible images. In the first phase of development, we focused primarily on synthetic data generated by the Astos camera simulator\footnote{\url{https://www.astos.de/products/camsim}.} and limited our pose estimate to a single degree of freedom. We trained two models based on Resnet18 \cite{he2016deep} and Resnet34 on visible images of our considered target spacecraft, Envisat.

From our results on our synthetic images, we were able to demonstrate that Resnet-based architectures trained on visible images can be robust on \gls{lwir} images. We also found that the deeper networks were able to achieve higher F1 scores. Work by \textcite{hajebi2008structure} has found that thermal infrared has its own unique challenges. These include noise and low resolution, reflections, halo effects or saturation, and history effects due to the time it takes temperature variations to propagate. These effects are difficult to model even when the thermal signature of the target is properly known. Hence, moving forward into the second stage of our phase of our development, we sought to demonstrate our model’s robustness in going from synthetic to real imagery.

The aforementioned works on the use of \glspl{cnn} for this application have failed to recognise they are complex and unintuitive tools, making them hard to trust in critical applications. This mistrust is based on the difficulty in understanding the origin source of failure cases and the consequent obstacles in debugging. This issue is not limited to our application: e.g.,\ there is much concern over the use of \glspl{dnn} in general for Earth-bound autonomous vehicles \cite{he2021explainable}. In this study we investigate our network further to attempt to explain reasons for failure using Grad-CAM and better understand what it focuses on to make good predictions \cite{selvajaru2019gradcam}.

Grad-CAM allows us to explore which regions of the image, and thus which parts of the target, were most important to the model in making its decisions. Any image processing technique that utilises model-based pose estimation must consider the possibility that the non-cooperative target may be damaged, may contain movable sections, or that the model might contain incorrect dimensions or be missing features. Thus, the knowledge provided by Grad-CAM can assist in mission planning since we can consider whether a network’s reliance on certain features may be inappropriate or not.

\section{METHODOLOGY}

\subsection{Viewsphere Sampling}
\label{sec:method-ssec:viewsphere}

The objective of our algorithm is to provide an initial estimate of the relative orientation of the target based on its appearance. This information can then be passed to more precise pose estimation algorithms, or be used to cross-reference with another sensor reading, or algorithm prediction. In our study, we are not attempting to predict the roll angle of the camera, nor the position of the target relative to the chaser; instead, we are interested in recovering the viewpoint under which the target is being imaged by the chaser, modelled as a classification problem.

\begin{figure}[t]
\centering
    \includegraphics[width=8cm]{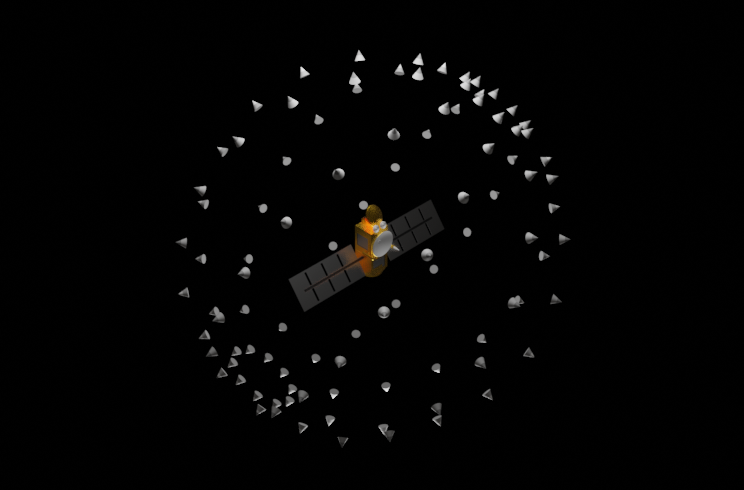}
        \caption{Illustration of a viewsphere around our target spacecraft, Jason-1 (not to scale).}
        \label{fig:viewsphere}
\end{figure}

We utilise the same concept as \textcite{rondao2021robust} of viewsphere sampling which is taken from 3D object detection. This concept imagines a sphere around the target, the surface of which is covered in cameras pointed toward the target at the centre, as can be seen in \figref{fig:viewsphere}. Any viewpoint can be described using the spherical coordinates of a camera on this sphere. 

During the first phase of our development we focused on one \gls{dof}, therefore the labels only corresponded to the azimuth. We used 10 classes, so that each class would represent a range of facets within 18 degrees from the centre of that class. The images in this dataset depicted the target spacecraft in a continuous roll, therefore there is variance in the observed facet within the classes.

During the second phase, we wanted to extend the pose estimation to two \glspl{dof}. Hence, each image was given two labels, one for the azimuth, and one for the elevation. We generated images for discrete steps of 10 degrees which were at the centre of our classes, therefore there is no variance in the observed facet within the class. Our new class scheme meant that each class represented a range of facets with 5 degrees from the centre of the class.

\subsection{Network Architecture}

\glspl{cnn} are specifically suited for multidimensional data such as images \cite{lecun1989handwritten}. They work on optimising convolutional kernels that are used to extract features from images. Deeper networks with more layers can extract more complicated features. However, deeper networks are also more prone to overfitting, something which can be problematic for our application since our test set is a different modality to our training set. We want our network to be good at generalising the shapes of the object since the textures and shades will appear different between the visible and thermal modalities. 

In order to overcome overfitting, the \glspl{cnn} used in this study are adapted versions of ResNet. This network has demonstrated good generalisation from winning many competitions in image recognition and classifications \cite{he2016deep}. It uses skip connections between layers in order to avoid the issue of vanishing gradients which occurs when training networks using gradient descent algorithms. In order to assist in training, ResNet utilises batch normalisation right after a convolution layer, and before an activation layer, this helps the gradient stay consistent, and not get so large that they slow down the network, or prevent it from training. 

Unlike visible images, thermal images only have one colour channel that represents the heat radiance. Therefore, the first layer of the model was adapted to accept an image with a depth of 1. In our implementation we use a focus module, which was inspired from the first layer of YoloV5 \cite{glenn_jocher_2021_4679653}. The focus module transfers spatial information to the channel dimension, allowing us to use higher resolution input images without needing to adapt any other layers of the network, or increase the depth of the network. The benefit of a higher resolution image is that the features of the target are maintained, an important benefit when running inference on \gls{lwir} images which are typically lower resolution than visible images.

The final layer of the \glspl{cnn} was adapted to suit our labels; in the first phase we adapted the number of outputs to 10, one for each class. This was changed again in the second phase when we shifted to predicting to two degrees of freedom. A single fully connected layer to predict would significantly reduce the number of samples available per class, whilst unnecessarily increasing the complexity of the model with a single layer with 648 possible classes. Therefore, two linear layers were added in parallel, one for each degree of freedom, one with 36 outputs for azimuth and the other with 18 for elevation.

\subsection{Envisat Dataset}

During the first phase of our development, we used a synthetic dataset generated by the Astos camera simulator featuring Envisat, a decommissioned oceanography satellite. Envisat was launched in 2002 and operated for over ten years until its mission was concluded in 2012 when communication was lost; it is currently adrift in near-Earth polar orbit.

The dataset contains images of Envisat, in both visible and \gls{lwir} modalities, arranged in sequences that simulate certain scenarios that might be faced by a chaser spacecraft as it approaches. The dataset has been previously used to assess classical feature detection and description algorithms in the context of space rendezvous; see \textcite{rondao2020benchmarking} for details on the data generation for both modalities.

\begin{figure}[t]
  	\centering
	\begin{subfigure}{0.49\textwidth}	\centering
		\includegraphics[trim=0 100 0 100, clip, height=0.2\textheight]{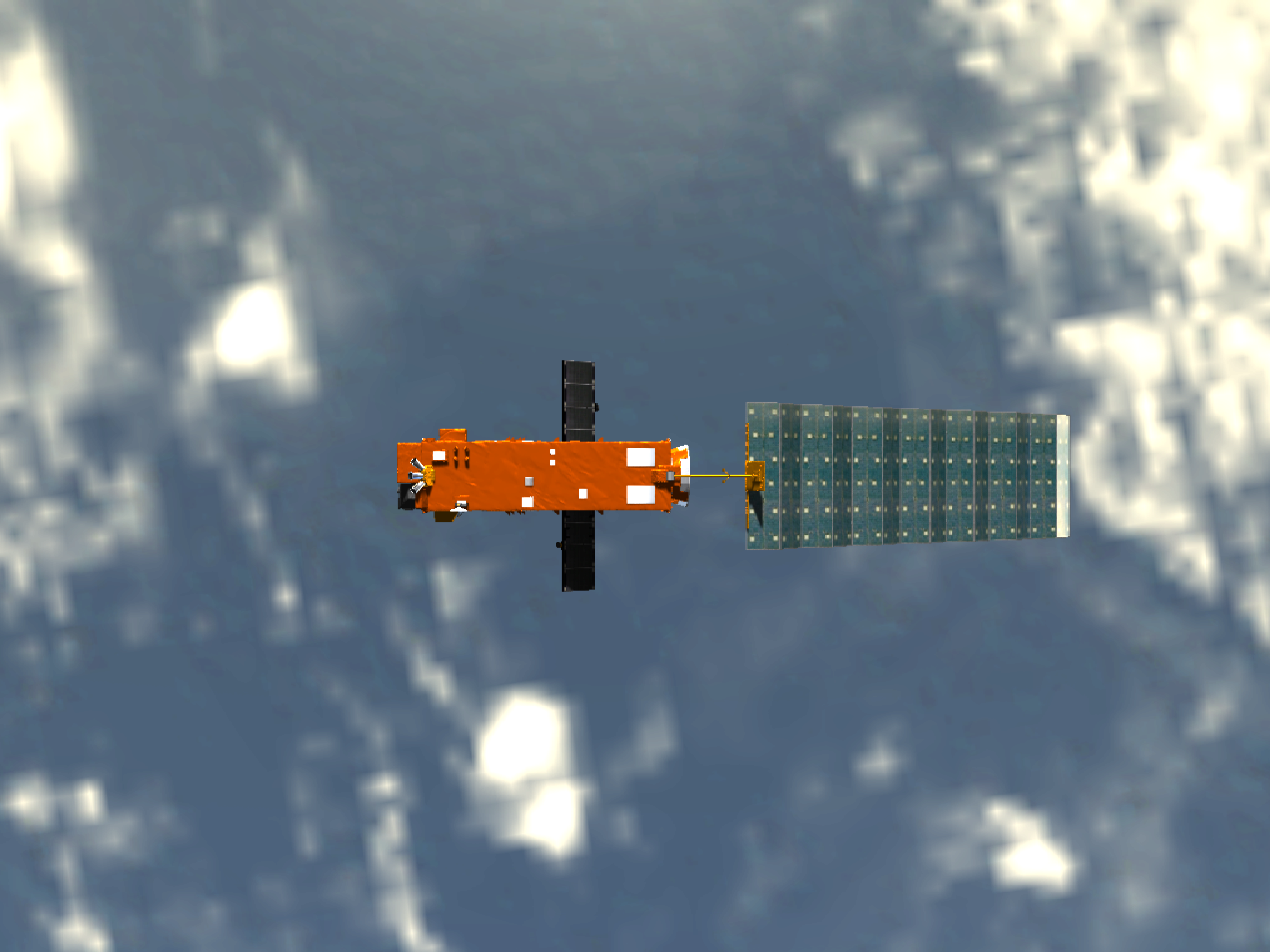}
		\caption{Visible}
		\label{fig:2vis}
	\end{subfigure}
	\begin{subfigure}{0.49\textwidth}	\centering
		\includegraphics[trim=0 100 0 100, clip, height=0.2\textheight]{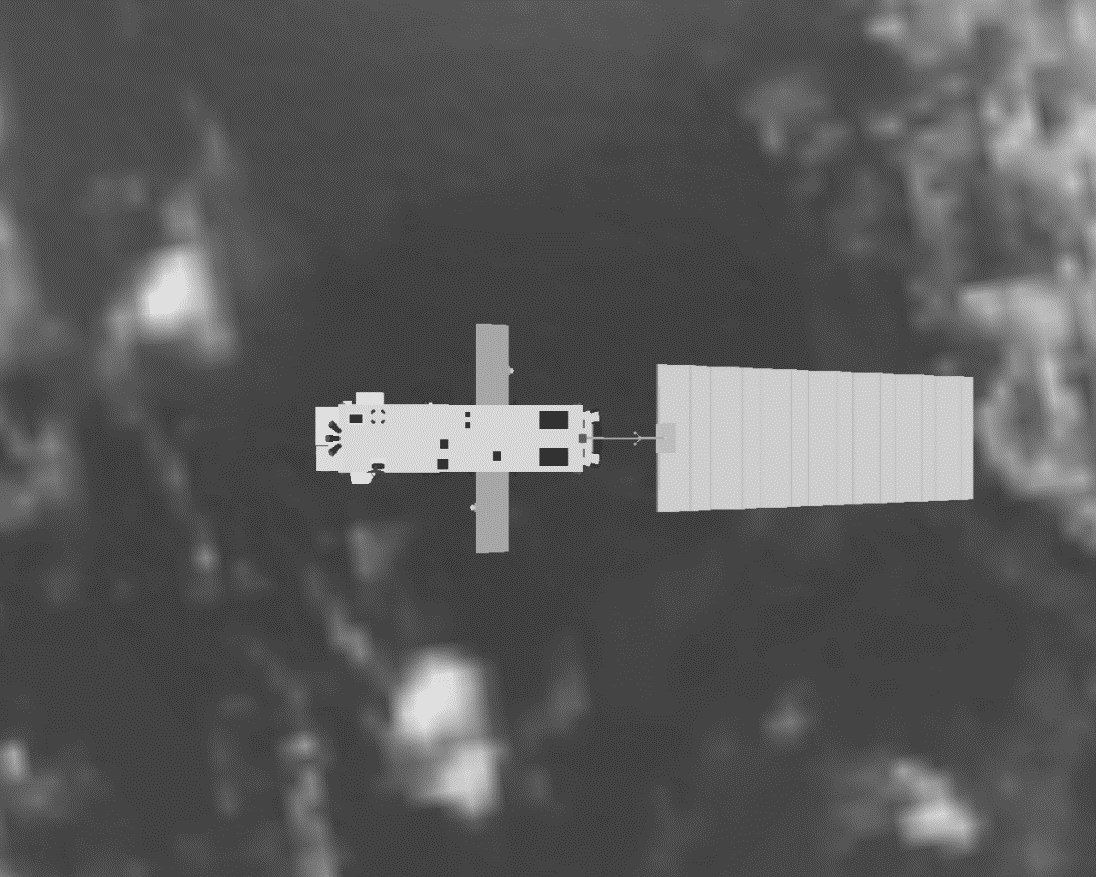}
		\caption{Thermal infrared}
		\label{fig:2ir}
	\end{subfigure}
\caption{Example images from the R-bar approach in the Astos Dataset featuring the Envisat spacecraft.}
\end{figure}

For the present study, we consider a dataset formed through a \SIrange{50}{50}{\percent} sampling from two simulated trajectories: 
\begin{enumerate*}[label=\arabic*)]
\item a negative V-bar approach vector such that the target is imaged against a black, deep space background; and
\item a negative R-bar approach, such that the target is imaged against the earth.
\end{enumerate*}
In both cases, the chaser is at a hold point at \SI{50}{\metre} relative distance, and the target is spinning about a single axis. \Figref{fig:2vis} shows an example of one such image from the R-bar sampling with the earth in the background, whereas \figref{fig:2ir} shows the same image in the \gls{lwir}.

Initially, a test was devised to train modified versions of Resnet18 and Resnet34 to be able to classify facets of the target into 10 classes, which each represent \SI{36}{\degree} of rotation along the azimuth. The network was trained purely on visible images, with each class containing 408. This included an even split of images taken from the R-bar and V-bar approach, and an even split of images showing the target in direct sunlight an eclipse. A subset of \SI{20}{\percent} of the training set was set aside as a validation set which could be used to stop training early to avoid over fitting. The test set contained the thermal counterparts of the visible images in the training set.

Both the visible and the thermal images are stored in \texttt{.PNG} format; as such, both have three colour channels in their current state. Whereas in the visible each channel represents the response in the red, green, and blue wavelengths separately, the thermal infrared represents a linear combination of the radiance for three sample wavelengths within the \gls{lwir} and the sensitivity of the simulated camera into a single image, and therefore the colour channels are all equal.

\subsection{Jason-1 Dataset}
\label{sec:method-ssec:jason1}

Our chosen target for the second phase was Jason-1, another decommissioned oceanography satellite, which is currently in low Earth orbit around the equator. Jason-1 was a joint venture between \gls{nasa} and \gls{cnes}, and was originally launched in December 2001. In early 2012, Jason-1 was retired and placed into a graveyard orbit and contact was lost in June of that year. The satellite is not expected to re-enter Earth’s atmosphere for at least 1000 years; for this reason, as well as its in-operation, it is a prime candidate for an \gls{adr} mission \cite{Jason1_blog}.

\begin{figure}[t]
  	\centering
	\begin{subfigure}[h]{0.49\textwidth}	\centering
		\includegraphics[width=\textwidth]{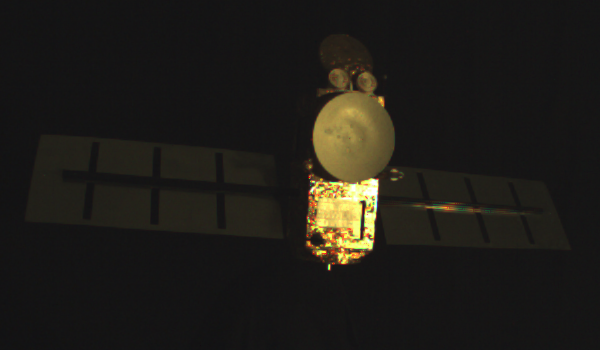}
		\caption{Visible}
		\label{fig:3a}
	\end{subfigure}\hfill
	\begin{subfigure}[h]{0.49\textwidth}	\centering
		\includegraphics[width = \textwidth]{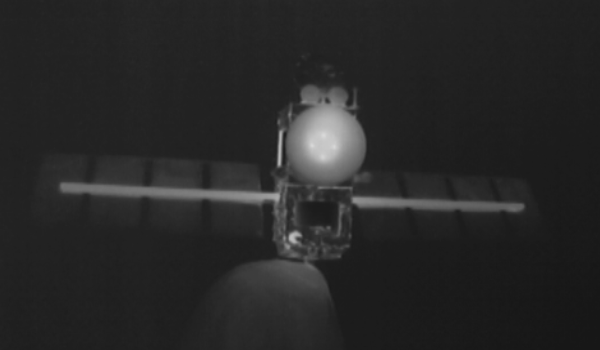}
		\caption{Thermal infrared}
		\label{fig:3b}
	\end{subfigure}
\caption{Example images from the Jason-1 Dataset.}
\label{fig:fig3}
\end{figure}

Utilising a free to use \gls{cad} model of Jason-1 as a basis, a laboratory mock-up was created, with a 1:4 scale replica with 1-\gls{dof} in rotation. \Figref{fig:fig3} shows images collected of the mock-up in the lab in the visible and thermal infrared modalities. Whilst we tried to make our replica as faithful as possible to the real counterpart, some smaller components are not included since they would not be able to support themselves with their new scale, and other components needed to be reinforced such as the radiometer and the solar panels. 

We did not have access to information on the thermal signature of Jason-1 in order to choose the materials for our replica. Therefore, we had to postulate from the data we had on Envisat (see \textcite{rondao2020benchmarking,yilmaz2018infrared} for details) and heated the mock-up manually to achieve a similar signature. Both crafts share many of the same components (e.g., insulating foil, radiators); however, the attained thermal signature on Jason-1 is only an approximation since in reality they share different orbits and hence different Sun exposure patterns.

For training, we used the mechanical \gls{cad} model of the replica to generate synthetic visible images using the free and open source Blender modelling software.\footnote{\url{https://www.blender.org}.} The solar arrays of the \gls{cad} model were textured with real images from the target, whereas the foil was procedurally generated. Using the viewsphere sampling method outlined before (see Sec.~\ref{sec:method-ssec:viewsphere}), we generated all the possible facets of the 3D target at steps of \SI{10}{\degree}. Since we were working in two \glspl{dof}, we generated the facets across the entire \SI{360}{\degree} of azimuth and \SI{180}{\degree} of elevation, covering the complete viewsphere. Additionally, we created 21 copies of each image with different combinations of lighting conditions to create reflections and shadows on the body.

As stated before, our network uses two fully connected layers in parallel, one to produce a class for azimuth, and one to predict a class for elevation. Therefore, images belonging to any one of the 36 azimuth classes could belong to the same class for elevation, and conversely an image belonging to a single azimuth class could belong to any of the 18 classes representing elevation. Considering we have 21 copies of each image, this means that each class of azimuth had 378 training images, and each class of elevation had 756 training images.

For testing, we acquired a continuous sequence of images according to the laboratory setup in the next section. The replica Jason-1 rotates at \SI{6}{\degree} per second, and we captured two full rotations. Our cameras were acquiring at \SI{10}{\hertz}, which would give us around 1200 images, in reality we captured 1174 images for the test set.

\subsection{Laboratory Setup}

\begin{figure}[t]
  	\centering
	\begin{subfigure}[h]{0.49\textwidth}	\centering
		\includegraphics[width=\textwidth]{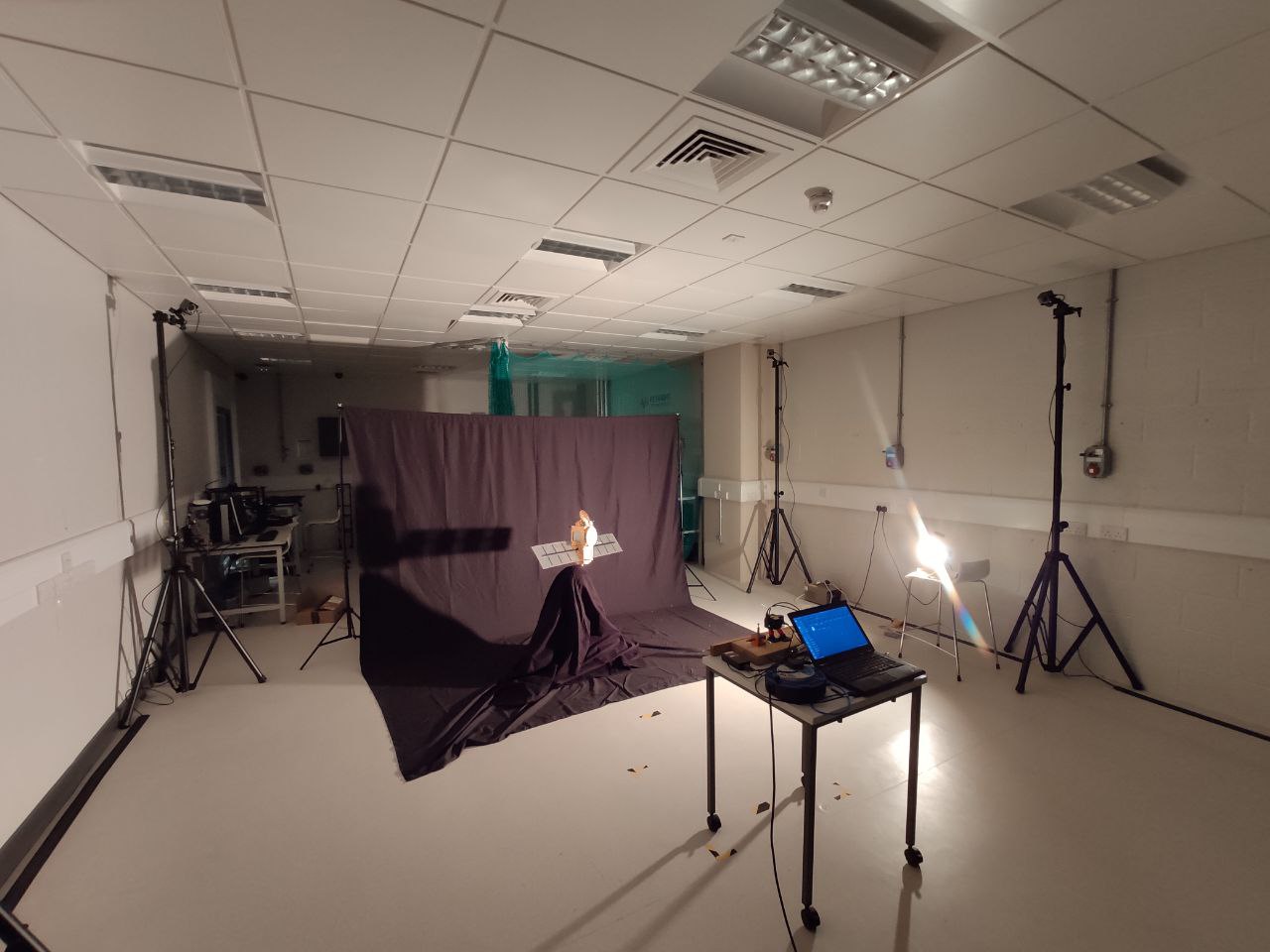}
		\caption{Equipment layout}
		\label{fig:lab}
	\end{subfigure}\hfill
	\begin{subfigure}[h]{0.49\textwidth}	\centering
		\includegraphics[width = \textwidth]{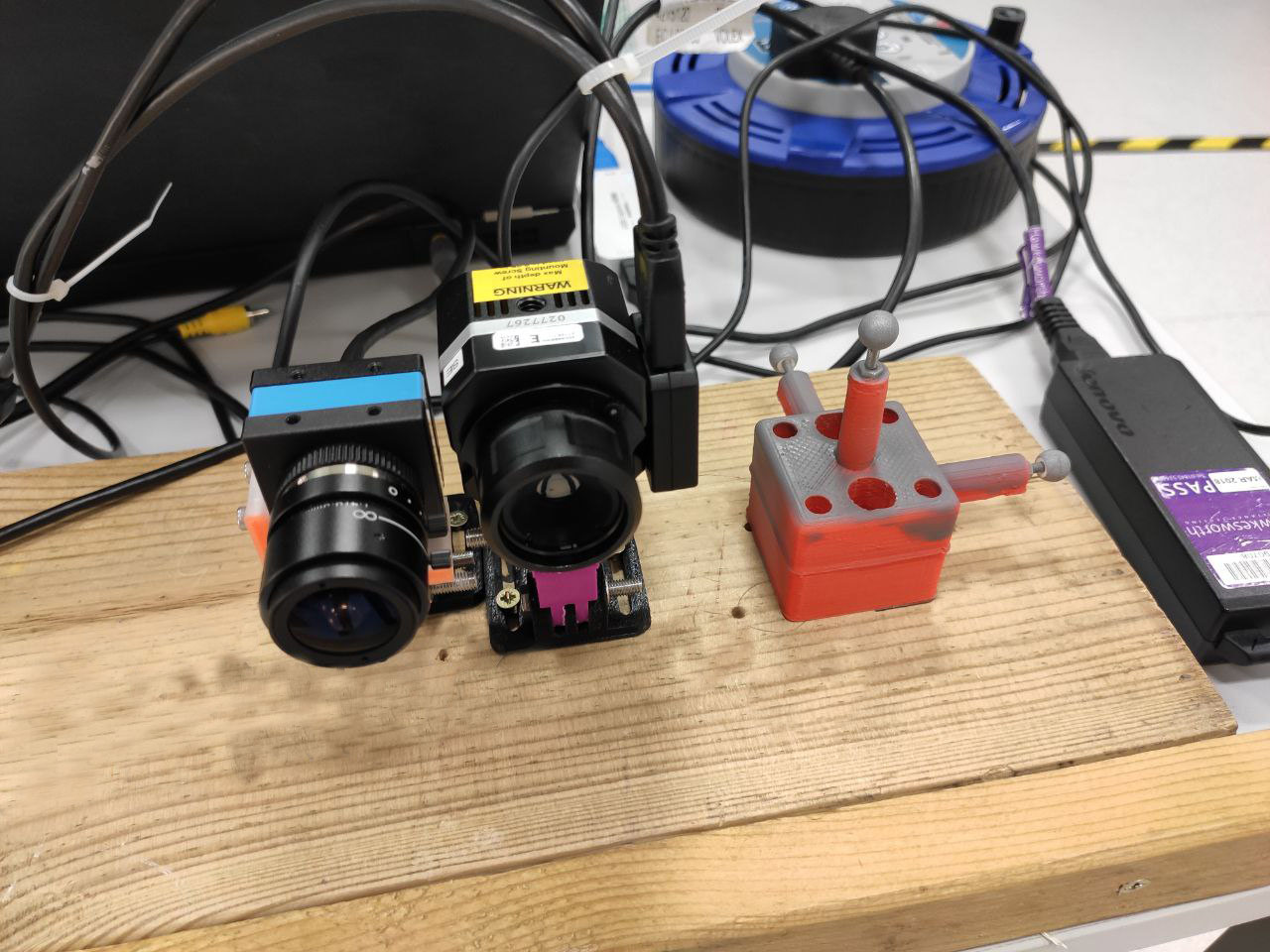}
		\caption{Multimodal camera configuration}
		\label{fig:camera}
	\end{subfigure}
\caption{\gls{asl} setup at City, University of London, for the acquisition of real images.}
\label{fig:asl}
\end{figure}

Data was collected in the \gls{asl} at City, University of London. An image of the target mock-up within our laboratory setup can be seen in \figref{fig:lab}. A blackout curtain was placed behind the target to simulate a deep space background. A floodlight was used to illuminate the target. 

The facility is equipped with an OptiTrack motion tracking system that can record the 6-\gls{dof} of a target by detecting, tracking and triangulating markers that are placed in the target. The OptiTrack system was used to track the rotation of the target to provide the ground truth. We captured the target rotating around its vertical axis at a fixed distance of \SI{1}{\metre}. 

Our camera setup can be seen in \figref{fig:camera}, with which both visible and thermal infrared images were captured simultaneously. This allowed us to compare the performance of our network on the real visible and real \gls{lwir}. The cameras are mounted side to side with a very small baseline to minimise the discrepancy between the images.

\subsection{Augmentation Techniques}

\begin{figure}[t!]
\centering
    \includegraphics[width=\textwidth]{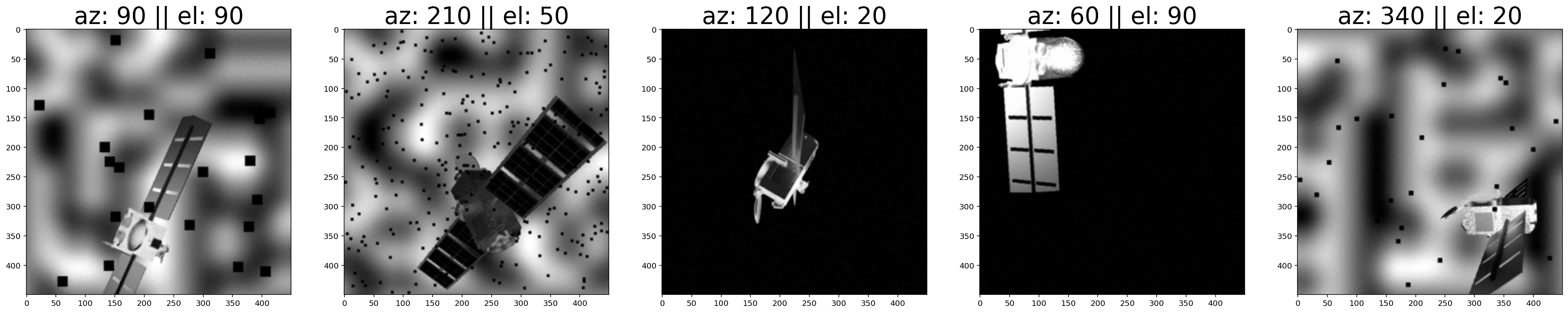}
        \caption{Montage of images exemplifying some of the transformations introduced by our data augmentation pipeline.
    \label{fig:fig4}
}
\end{figure}

Image augmentation is a powerful tool which can be used to generate more data for which the network can learn from. However, it must be representative of the data that the developer is intending to encounter in the real-world when deploying the model. Carelessness could result in unwanted biases introduced into the dataset. In the first phase, we used augmentations including colour augmentation to randomly change the brightness and hue before converting to greyscale. This would create new images of Envisat with different shades which would mimic what we may expect from thermal images. 

We also applied positional and scale augmentations to randomly shift the image around the frame and scale up or down the image. When we moved to the second stage, we increased the magnitude of the translation and scale augmentations in order to make the network more robust to the variance of the target’s location within the frame. We took advantage of a technique presented by \textcite{proenca2019deep} which used homography augmentations to warp the image in such a way to simulate variance in the camera rotation and roll. 

We are concerned about our network’s robustness to variance in the thermal signature. As such, we wanted to prevent the network from becoming too dependent on certain shades of different components. The thermal signature could change depending on the time that the target had been in direct sunlight and for how long. Therefore, when we moved into the second stage of development we began to train on the red and blue channels of the synthetic images and validate on the green channel. Moreover, we randomly augmented the gain and contrast of the training set.

Our images were generated with an alpha channel so that we could easily mask the target and modify the background prior to it being loaded into the model. In order to simulate a diverse background, Perlin noise was generated and inserted into the background via the alpha channel. Perlin noise is typically employed by visual effects artists to increase the appearance of realism in computer graphics. In addition, we introduced additive Gaussian noise across the image to simulate sensor noise, and employed Gaussian blur so that features would appear softer, similar to what may be seen in thermal images.

Like \textcite{proenca2019deep}, we also employed patch dropout to randomly remove super pixels from the image, thus occluding certain features. This would encourage the network to rely on a more diverse group of features in order to make its predictions, making it more robust to differences between the computer model and the replica. \Figref{fig:fig4} shows examples images after our augmentations had been applied.

\section{RESULTS}

\subsection{Evaluation on Synthetic Dataset}

\begin{table}[t!]
    \begin{center}
        \caption{Performance metrics of our Resnet18 and Resnet34 architectures on the Envisat test set composed of synthetic thermal infrared images.}
        \label{tab:envisat}

        \begin{tabular}{ |c|c|c|c| } 
        \hline
        {\textbf{Architecture}} & {\textbf{Precision}} & 
        {\textbf{Recall}} & {\textbf{F1 Score}} \\ \hline
        
        Resnet18               &   0.819    &   0.839   &	0.819 \\ \hline
        Resnet34               &   0.883    &   0.896   &   0.883 \\ \hline

        \end{tabular}
    \end{center}
\end{table}

\begin{figure}[t!]
  	\centering
	\begin{subfigure}[t]{0.5\textwidth}	\centering
		\includegraphics[width=\textwidth]{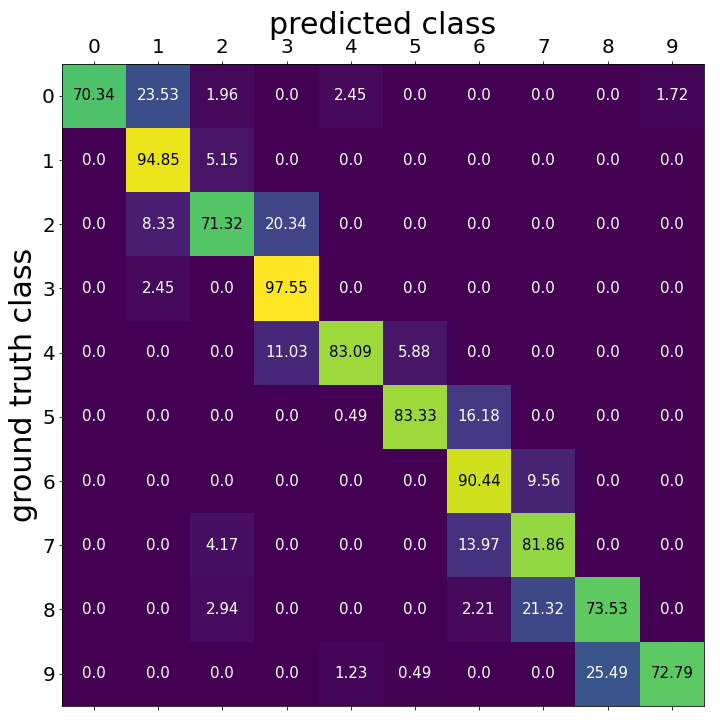}
		\caption{Resnet18}
	\end{subfigure}\hfill
	\begin{subfigure}[t]{0.5\textwidth}	\centering
		\includegraphics[width = \textwidth]{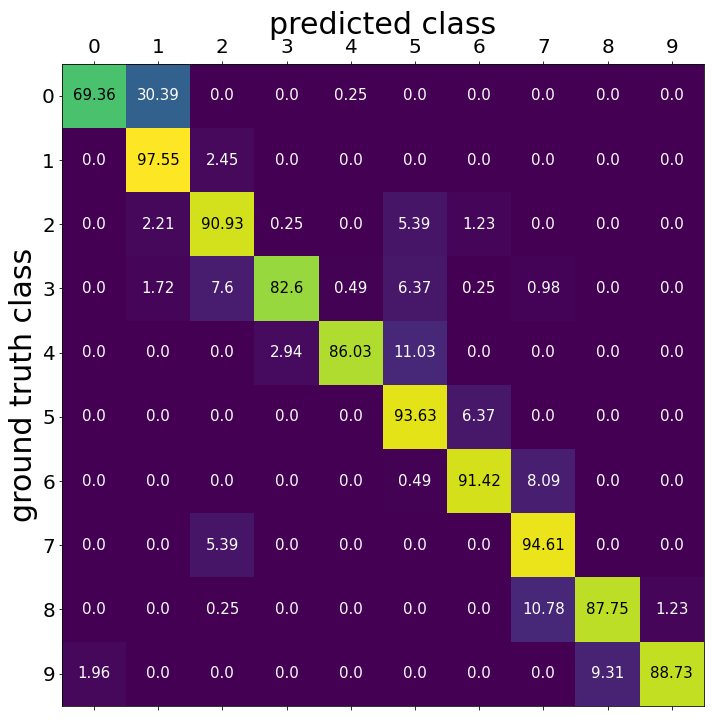}
		\caption{Resnet34}
	\end{subfigure}
\caption{Confusion matrices of results of our models on the Envisat test set composed of synthetic thermal infrared images.}
\label{fig:fig5}
\end{figure}

As stated before, in the first phase of development, we focused on proving the ResNet’s architecture suitability for our application. In order to do so we had to demonstrate that it was capable of classifying previously unseen \gls{lwir} facets of our target. For this purpose, the precision, recall, and F1 score metrics are used; these are commonly employed to gauge the performance of a classification algorithm. Precision can be calculated using \eqref{equ:precision}, and is a measure of how many of the predictions allocated to a given class were in fact correct. Recall is calculated using \eqref{equ:recall}; as can be seen this metric accounts for false positives that occur in a given class. 

\begin{align}
    \text{precision} &= \frac{\text{true positives}}{\text{true positives} + \text{false positives}}\label{equ:precision}\\
    \text{recall} &= \frac{\text{true positives}}{\text{true positives} + \text{false negatives}}\label{equ:recall}
\end{align}

F1 score takes into account both precision and recall, and is calculated with \eqref{equ:f1}. Given that false negatives and false positives are equally unwanted in our application, the F1 score will be an appropriate global metric to judge our network.

\begin{equation}
    \text{F1}_\text{score} = 2 \times \frac{\text{precision} \times \text{recall}}{\text{precision} + \text{recall}}
    \label{equ:f1}
\end{equation}

During this investigation we trained two models based on Resnet18 and Resnet34 architectures. \Tabref{tab:envisat} shows the average precision, average recall, and average F1 score across all the classes. It can be seen that the deeper network did perform better on the test set in all three metrics. 

These metrics do not measure the distance between the predicted class and the ground truth. In our application, a misclassification towards an adjacent class is not a major issue as it only represents a small rotation error. Hence, we also created the confusion matrices to look at where false predictions occurred; these can be observed in \figref{fig:fig5}. One can see that the majority of false positives were misclassified as the adjacent class. 

Both these results from \Tabref{tab:envisat} and \figref{fig:fig5} indicate that the networks we developed both performed well on the task presented to them. On the other hand, this represented a simpler problem as only one \gls{dof} was considered. In addition, a relatively small number of classes (10) was considered, which meant that each one contained a relatively large set of viewpoints. This allows more leeway for the model's predictions.
\subsection{Evaluation on Real Dataset}

\begin{figure}[t!]
  	\centering
	\begin{subfigure}[h]{0.495\textwidth}	\centering
		\includegraphics[width=\textwidth]{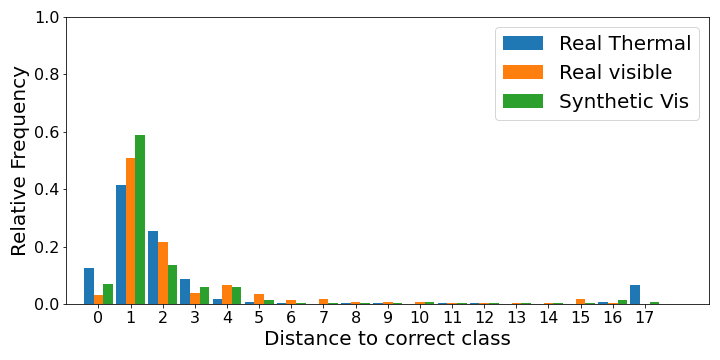}
		\caption{Azimuth error}
	\end{subfigure}
	\begin{subfigure}[h]{0.495\textwidth}	\centering
		\includegraphics[width = \textwidth]{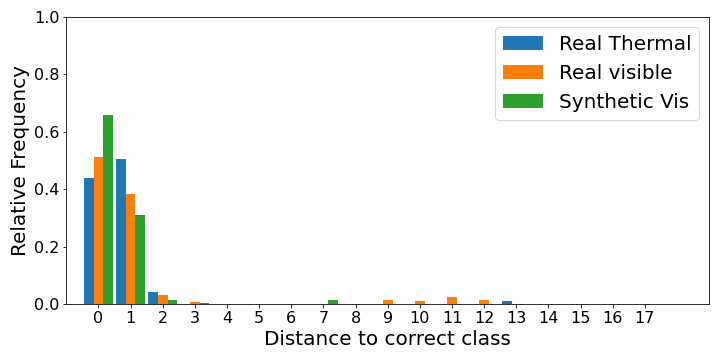}
		\caption{Elevation error}
	\end{subfigure}
\caption{Histogram of results on the Jason-1 real visible, real thermal infrared, and synthetic visible trajectory test sets.}
\label{fig:fig6}
\end{figure}

\begin{table}[t!]
    \begin{center}
        \caption{Absolute angular errors $\theta_\text{error}$ between the predicted viewpoint and the ground truth viewpoint.}
        \label{tab:error}

        \begin{tabular}{ |c|c|c| } 
         \hline
        \multirow{2}{*}{\textbf{Condition}} & \multicolumn{2}{c|}{\textbf{Error (\si{\degree})}}\\ \cline{2-3}
                                        &  \textbf{Mean} & \textbf{Median} \\ \hline
        Synthetic Visible               &   12.39      &	 8.34 \\ \hline
        Real Visible                     &   18.23      &    10.15 \\ \hline
        Real Thermal                     &   19.27      &	10.75 \\ \hline
        \end{tabular}
    \end{center}
\end{table}

As mentioned in Section \ref{sec:method-ssec:jason1}, the Jason-1 training set was generated at discrete \SI{10}{\degree} steps in azimuth and elevation. However, the test set consists of a continuous trajectory which includes viewpoints in between those steps, thus evaluating the robustness of the model towards unseen perspectives. In addition to this, we created an additional test set by recreating this trajectory inside the Blender simulation environment by importing the ground truth recorded by Optitrack, along with the same lighting conditions. This provides a baseline test set which we could then use to compare the performance of our algorithm between synthetic and real data. 

The orientation of the target in the test sets was grouped into bins of \SI{10}{\degree} in order to measure the distance between the predicted class and the ground truth class labels. \Figref{fig:fig6} shows the results for the performance on both the azimuth and elevation. Each colour represents a different condition: green for the synthetic visible test set; orange for the real visible test set; and finally blue for the real thermal infrared test set. For all three conditions, the majority of the predicted classes are found on the left-hand side of the two graphs, indicating that the network performed well on both \glspl{dof}. 

The network performed better on the task of predicting the elevation --- in all three conditions, over \SI{90}{\percent} of the predictions were within a 1 bin distance, which is equivalent in a worst case scenario to an error of \SI{20}{\degree}. In the case of the azimuth, only \SI{60}{\percent} of the synthetic set and \SI{50}{\percent} of the real sets were within 1 bin. Nevertheless, we did find that around  \SI{80}{\percent} of the predictions for both real sets were within 2 bins, and thus within \SI{30}{\degree} (worst case scenario) from the ground truth. 

We also calculated the absolute error between the predicted facet and the ground truth using \eqref{equ:error}. To do so, we consider an imaginary line-of-sight vector drawn from the camera to the centre of the target. For the ground truth we denote this vector as $V_\text{gt}$; similarly for the prediction we use $V_\text{pr}$. Since we are not considering the distance between the target and the chaser we can treat this as a unit vector. We can, therefore, take the inverse cosine of the dot product between the $V_\text{gt}$ and $V_\text{pr}$ to represent the principal angle error (disregarding the camera roll, since the viewpoint does not change with it):

\begin{equation}
    \theta_\text{error} = \arccos{(V_\text{gt} \cdot V_\text{pr})}
    \label{equ:error}
\end{equation}

\Tabref{tab:error} shows the mean and median $\theta_\text{error}$ measured across the entire trajectory. It shows that for all three conditions the average error was less than \SI{20}{\degree}. We also found that the average error for both the lab datasets were close. It can be seen that approximately \SI{10}{\percent} of the predictions on the real thermal images are located \SI{180}{\degree} away from the ground truth, as indicated by the spike on the right-hand side of the plot in \figref{fig:fig6}. This would contribute significantly to the average absolute error. We speculate that the cause of this phenomenon is due to the symmetrical nature of our target.

\subsection{Evaluation Using Grad-CAM}

\glspl{dnn} have achieved unprecedented performance in many applications, although they remain opaque tools. Simply inspecting the model’s structure, or analysing its weights, tells us very little about its decision-making behaviour. In addition, relying on simple metrics such as F1, recall and precision, can lead a developer to overestimate the capability of their models. In order to better understand the models developed on Jason-1, we have employed Grad-CAM to identify the sections of the target that the model relied on to make predictions.  

This could be done in different ways, either by encoding both predictions for azimuth and elevation, or we could treat them as mutually exclusive, and investigate one or the other. Initially we were interested in investigating where the network’s focus was considering both, since this will give us a general sense of how it made decisions. This is done in much the same way as outlined by \textcite{selvajaru2019gradcam}, except since we have two outputs, each is one-hot encoded, and then summed together prior to backpropagation. 

\begin{figure}[t]
  	\centering
	\begin{subfigure}[h]{0.3\textwidth}	\centering
		\includegraphics[width=\textwidth]{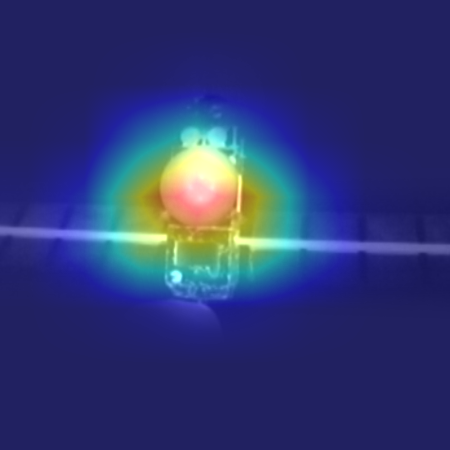}
	\end{subfigure}
	\begin{subfigure}[h]{0.3\textwidth}	\centering
		\includegraphics[width = \textwidth]{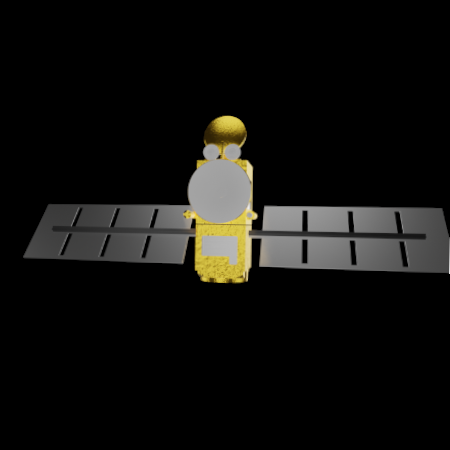}
	\end{subfigure}
\caption{Illustration of Grad-CAM for \gls{dnn}-based spacecraft pose estimation explainability tested on the \gls{lwir} modality. The left figure shows the thermal infrared image which was fed to the network, overlaid with the Grad-CAM attention map generated by considering both the azimuth and elevation. The right figure shows the closest match according to the network.}
\label{fig:frame_30}
\end{figure}

We discovered that, throughout the trajectory, the network’s attention was focused heavily on the central bus; an example is shown in \figref{fig:frame_30}. This is logical since it is both the centre of the target and also where many of its unique components are found, such as the radiators and instrument antennas. These are the components which would be absent if we were looking at the target from a vastly different orientation.

\begin{figure}[t]
  	\centering
	\begin{subfigure}[h]{0.3\textwidth}	\centering
		\includegraphics[width=\textwidth]{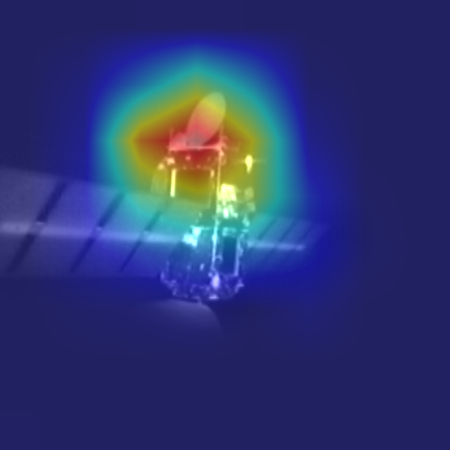}
	\end{subfigure}
	\begin{subfigure}[h]{0.3\textwidth}	\centering
		\includegraphics[width = \textwidth]{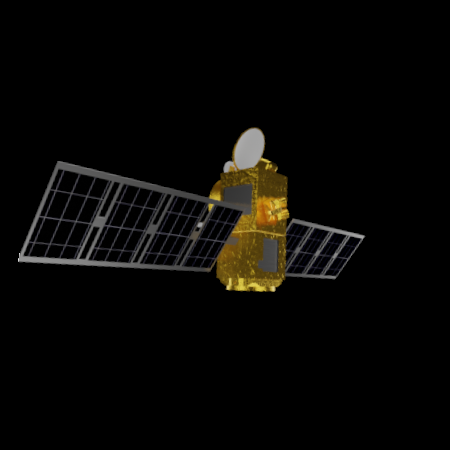}
	\end{subfigure}
\caption{Illustration of a good match between the ground truth (left) and the prediction (right). The Grad-CAM was applied only taking into account the azimuth class.}
\label{fig:frame_214}
\end{figure}

We were also interested in what caused the network to misclassify the azimuth of some thermal infrared images \SI{180}{\degree} away from the ground truth. Therefore, we applied Grad-CAM again, but this time only using the one-hot encoding of the azimuth class, and while completely ignoring the elevation prediction. \Figref{fig:frame_214} shows an example where this method is applied to a correctly predicted orientation. We can see that the focus is on the bus, but particularly concentrated at the top near what would be the radiometer.

\begin{figure}
  	\centering
	\begin{subfigure}[h]{0.3\textwidth}	\centering
		\includegraphics[width=\textwidth]{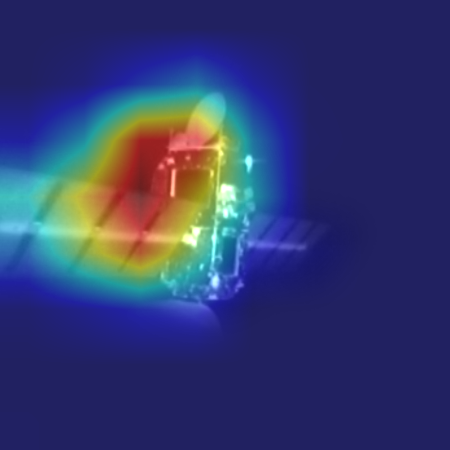}
	\end{subfigure}
	\begin{subfigure}[h]{0.3\textwidth}	\centering
		\includegraphics[width = \textwidth]{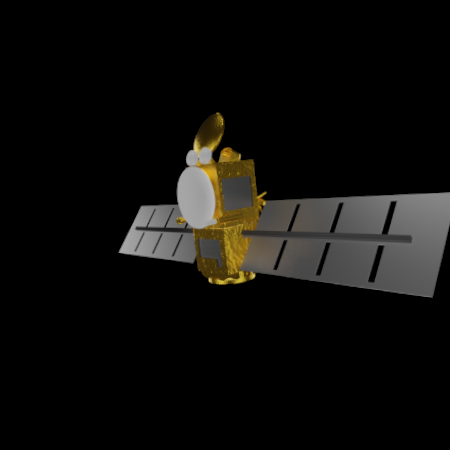}
	\end{subfigure}
\caption{Illustration of a bad match between the ground truth (left) and the prediction (right). The Grad-CAM was applied only taking into account the azimuth class.}
\label{fig:frame_215}
\end{figure}

Juxtaposed, \figref{fig:frame_215} shows the next frame which occurs in one of the instances where the model misclassified the target \SI{180}{\degree} out. We can see in this image the model’s focus has shifted away from the central bus to include the solar panels. This supports our speculation, stated in the previous section, that the symmetrical nature of the target caused the network to make the misclassification, since the solar panels are symmetrical. This is made more difficult in the thermal infrared, since the patterns on the front of the panels cannot be made out. Moreover, the supports for the panels radiated heat slower, and could be seen through the panels. 

\begin{figure}[t]
  	\centering
	\begin{subfigure}[h]{0.3\textwidth}	\centering
		\includegraphics[width=\textwidth]{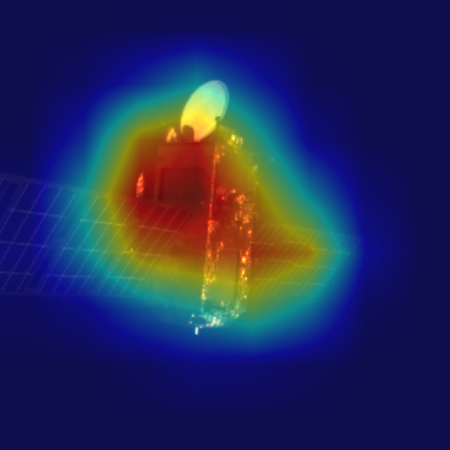}
	\end{subfigure}
	\begin{subfigure}[h]{0.3\textwidth}	\centering
		\includegraphics[width = \textwidth]{gradcam_imgs/vis214.png}
	\end{subfigure}
\caption{Illustration of Grad-CAM for \gls{dnn}-based spacecraft pose estimation explainability tested on the visible modality. The left image shows the visible image which was fed to the network, overlaid with the Grad-CAM attention map which was generated this only considering the azimuth. The right image shows the closest match according to the network.}
\label{fig:frame_vis}
\end{figure}

We investigated further by repeating this procedure on the real visible images we had collected. \Figref{fig:frame_vis} shows a frame from the visible camera that was captured at the same time as the offending frame shown in \figref{fig:frame_215} was captured. We found that, when viewing facets on this side of the target such as the one shown, much of the attention was on the surface of the solar panels. This indicates that the model is biased to search for this pattern, and the fact that it is absent from the thermal images put the network at a disadvantage. 

\section{CONCLUSION}

Our goal in this study was to develop a \gls{cnn} to predict the coarse viewpoint of a target as imaged by a rendezvousing chaser for the initialisation of spacecraft relative pose estimation algorithms, and to evaluate its robustness on thermal infrared imagery when it has been trained exclusively on visible images. During our investigation, we firstly demonstrated this capability on a synthetic dataset of the spacecraft Envisat. We then expanded our investigation to real images of the Jason-1 spacecraft captured in our laboratory. In this phase we demonstrated our network could perform just as well on real thermal infrared images as it could on real visible images.

Our experimental set up is currently limited to only one degree of rotation and does not allow us to test all possible permutations of azimuths and elevations. Therefore, we cannot verify our model’s performance on all possible orientations of a real target, which will need to be proven in order for our model to be deployed in the wild. Nevertheless, for the available facets of our real target, we have demonstrated our network is able to achieve an average accuracy greater than \SI{20}{\degree} in both modalities.

Moreover, we have gone a step further by attempting to explain how our network makes decisions. From our investigations with Grad-CAM we were able to better understand that the features on the central bus were key to the network's performance. Furthermore, we were able to better understand its failure modes, which we identified as the textures on the solar panels. This weakness can be removed from the \gls{cad} model in future work to achieve better performance in later iterations. 

As mentioned, the focus of our network tended towards the central bus of the target, therefore future work will examine how well the Grad-CAM maps match the target's location within the frame. From our observations in this study, we expect this can be used to explicitly predict a coarse position of the target. In addition, we will also investigate the performance of our network in other conditions which might occur on orbit such as in eclipse. This will allow us to evaluate in conditions where we'd expect thermal images to provide better clarity of the target than visible.

\printbibliography[title=REFERENCES]


\end{document}